\documentclass[letterpaper, 10 pt, conference]{ieeeconf}  

\IEEEoverridecommandlockouts                              

\usepackage{graphicx}
\usepackage{amsmath} 
\usepackage{amssymb}
\usepackage{multirow}
\usepackage{booktabs}
\usepackage[linesnumbered,ruled,vlined]{algorithm2e}
\usepackage{placeins}
\usepackage{float}
\usepackage[font=small]{caption} 
\usepackage[table,dvipsnames]{xcolor}

\title{\LARGE \bf
Non-differentiable Reward Optimization for Diffusion-based \\ Autonomous Motion Planning
}

\author{
Giwon Lee*$^{1}$, Daehee Park*$^{2}$, Jaewoo Jeong*$^{1}$, Kuk-Jin Yoon$^{1}$
\thanks{* denotes equal contribution}
\thanks{The authors are with (1) Department of Mechanical Engineering, KAIST and (2) Department of Electrical Engineering and Computer Science, DGIST. Email: \{dlrldnjs, jeong207, kjyoon\}@kaist.ac.kr and dhpark@dgist.ac.kr}%
}

\begin{document}

\maketitle
\thispagestyle{empty}
\pagestyle{empty}

\begin{abstract}
Safe and effective motion planning is crucial for autonomous robots. 
Diffusion models excel at capturing complex agent interactions, a fundamental aspect of decision-making in dynamic environments.  
Recent studies have successfully applied diffusion models to motion planning, demonstrating their competence in handling complex scenarios and accurately predicting multi-modal future trajectories.  
Despite their effectiveness, diffusion models have limitations in training objectives, as they approximate data distributions rather than explicitly capturing the underlying decision-making dynamics.
However, the crux of motion planning lies in non-differentiable downstream objectives, such as safety (collision avoidance) and effectiveness (goal-reaching), which conventional learning algorithms cannot directly optimize.  
In this paper, we propose a reinforcement learning-based training scheme for diffusion motion planning models, enabling them to effectively learn non-differentiable objectives that explicitly measure safety and effectiveness.  
Specifically, we introduce a reward-weighted dynamic thresholding algorithm to shape a dense reward signal, facilitating more effective training and outperforming models trained with differentiable objectives.  
State-of-the-art performance on pedestrian datasets (CrowdNav, ETH-UCY) compared to various baselines demonstrates the versatility of our approach for safe and effective motion planning.

\end{abstract}
\section{INTRODUCTION}

Motion planning is a pivotal component of the autonomy stack, following the perception and prediction of surrounding agents~\cite{hu2022st, sadat2020perceive, zeng2020dsdnet, jeong2025multi}. 
Its primary goal is to generate a safe and efficient trajectory for autonomous robots, leveraging environmental information obtained in preceding stages. 
Conventional approaches typically rely on analytical methods such as graph search~\cite{9981663, gammell2014informed, yonetani2021path, karaman2011sampling}, which can explicitly encode constraints like collision avoidance. 
However, they often struggle to handle complex scenes that involve intricate interactions and dynamic uncertainties.

\begin{figure}[t]
  \centering
   \includegraphics[width=\linewidth]{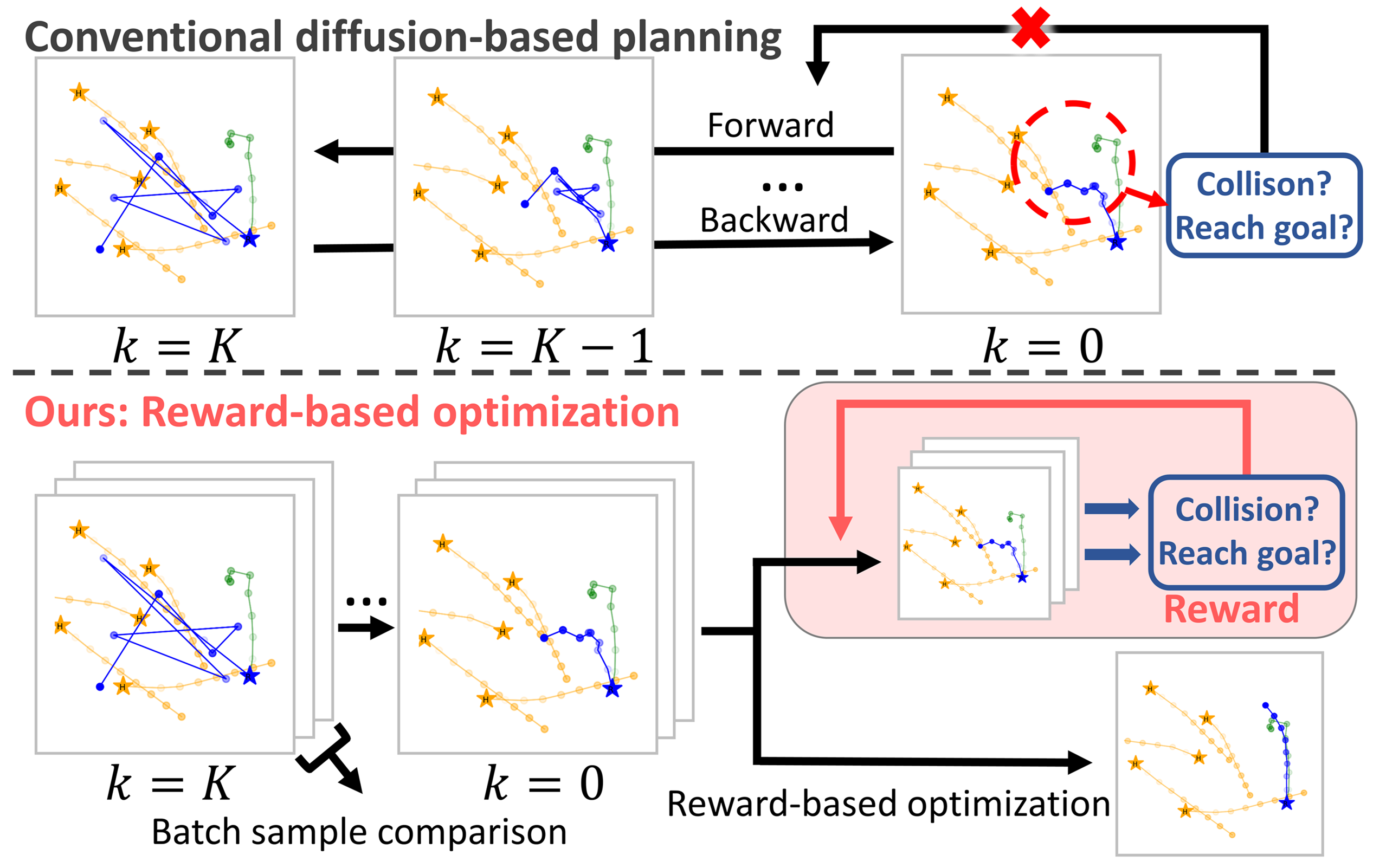}

   \caption{
   Conventional methods cannot train diffusion models on non-differentiable objectives such as collision rate or goal achievement. 
   In contrast, our method uses these objectives as rewards, explicitly training the diffusion motion planning model on key metrics.}
   \label{fig:viz_intro}
   \vspace{-5pt}
\end{figure}

Recently, deep learning has shown promise in addressing these challenges by handling high-dimensional inputs effectively~\cite{chen2019deep, zhang2021safe, pmlr-v229-dauner23a, diehl2022_review9}.
In particular, diffusion generative models have gained traction in motion planning, as they excel at modeling multi-modal future distributions in stochastic or uncertain scenarios~\cite{yang2024diffusion, jiang2023motiondiffuser}.
Nonetheless, a key limitation of current diffusion-based methods is that they are primarily trained via maximum likelihood estimation, which focuses on reconstructing the training data distribution rather than explicitly optimizing for critical downstream objectives, such as collision avoidance or goal success.
As illustrated in the upper row of Fig.~\ref{fig:viz_intro}, the sampling process in conventional diffusion-based methods is not guided by explicit planning metrics but instead learns to maximize the log-likelihood at each diffusion step.
Although gradient guidance~\cite{dhariwal2021diffusion, zhong2023language, chang2025safe} can help enforce certain constraints during sampling, this approach is only viable for differentiable objectives.
Critical objectives for planning related to safety and effectiveness are non-differentiable, which cannot be explicitly optimized with conventional training schemes.

For direct optimization on the non-differentiable metrics, we present a reinforcement learning (RL) approach that integrates non-differentiable reward signals into diffusion-based planners.
Recent advancements in RL have shown to significantly improve reasoning capabilities in diverse area including large language models (LLMs)~\cite{lang2024fine}.
We aim to leverage this capability to enable the model to reason and generate optimal plans that effectively satisfy non-differentiable objectives such as safety and effectiveness.

Our method is built upon Denoising Diffusion Policy Optimization (DDPO)~\cite{black2024training}, which formulates the diffusion generation process as a Markov Decision Process (MDP).
As shown in the lower row of Fig.~\ref{fig:viz_intro}, samples within a batch are compared based on a defined reward function, and the model is iteratively updated to maximize the reward.
However, policy optimization methods are known to struggle when rewards are sparse, meaning that only a small subset of generated samples satisfy a given condition.
Especially in autonomous motion planning, certain objectives (e.g., avoiding collisions or reaching the goal) may not frequently occur, leading to sparse reward signals. Therefore, directly applying the methodology of~\cite{black2024training} is not feasible.
To address this challenge, we propose a non-differentiable reward formulation with a dynamic thresholding algorithm that stabilizes learning and improves performance.
Specifically, as the rewards for samples within a batch become increasingly sparse, we adjust the threshold of non-differentiable rewards to extract meaningful reward signals from a larger number of samples, thereby addressing the sparse reward problem.
We validate our approach on diverse trajectory datasets and show that our method outperforms existing baselines in terms of safe and effective motion planning.
Furthermore, we demonstrate that our approach is not limited to predefined reward functions but can also optimize for other non-differentiable metrics, highlighting its flexibility in real-world planning scenarios.

In summary, our main contributions are as follows:
\begin{itemize}
    \item We propose a RL-based training framework for diffusion-based autonomous planning model to leverage the representational power of diffusion models while enabling optimization on non-differentiable downstream objectives.
    \item We address the sparse reward problem during training by introducing a dynamic thresholding algorithm that stabilizes training and improves overall performance.
    \item We conduct extensive experiments on diverse trajectory datasets, demonstrating the universality and state-of-the-art performance of the proposed method.
\end{itemize}
\section{RELATED WORKS}

\subsection{Generative Diffusion Model}
Generative diffusion models produce realistic and detailed samples through a progressive denoising process, widely applied in vision tasks~\cite{ho2020denoising, nichol2021improved, park2024improving}.
Diffusion models are typically trained by maximizing an approximation of the log-likelihood objective~\cite{song2021maximum}, upon which sampling results in effective multi-modal generation.
This is particularly valuable in autonomous systems, where modeling diverse future trajectories is crucial for safe and efficient planning~\cite{lu2024data, janner2022planning, parkleveraging}.
However, such approach is ineffective for real-world motion planning, as key metrics of safety and effectiveness are non-differentiable and could not be explicitly optimized.
This limitation underscores the need for methods incorporating task-specific objectives into the diffusion training process, generating realistic and deployable trajectory plans in safety-critical environments.

\subsection{Reinforcement Learning}
Reinforcement Learning (RL) is a framework where an agent learns to make decisions by optimizing a reward signal~\cite{sutton2018reinforcement}.
Unlike supervised learning, RL can also optimize non-differentiable rewards, making it advantageous on tasks where such rewards are the sole option~\cite{mnih2015human}.
A foundational framework in RL is the Markov Decision Process (MDP), which models decision-making as a sequence of state-action transitions~\cite{puterman2014markov, bellman1957markov}.
As a MDP, the agent observes the current state, selects an action, transitions to a new state, and receives a reward.
The goal is to learn a policy that maximizes the expected cumulative reward~\cite{silver2014deterministic}.
We optimize the discrete reward signals that are crucial for autonomous driving by modeling the diffusion model as MDP.


\subsection{Motion Planning for Autonomous Systems}
\label{subsec:motion_planning}
Autonomous planning aims for safe and efficient navigation, conventionally using optimization-based and sampling-based approaches.
Optimization-based methods generate trajectories by optimizing objectives under vehicle dynamics constraints but often struggle in highly dynamic environments~\cite{da2022comprehensive, chen2019autonomous}.
On the other hand, sampling-based methods offer flexibility in path generation but suffer from lack of smoothness and optimality~\cite{gammell2014informed, danesh2023leader, lei2021kb, chen2020driving}.
RL resolves these limitations by leveraging adaptive reward signals~\cite{chen2019crowd, zhang2022rethinking, diehl2021umbrella}. 
Generative models effectively learn these reward signals, capturing complex environmental dynamics and providing informative feedback for decision-making~\cite{kaiser2019model}.
We aim to apply such an advantage to trajectory planning by explicitly using key metrics as reward signals.

\section{METHODS}

\subsection{Problem Definition}
We consider a scene with the ego agent (index 0) and $N-1$ surrounding agents.  
The observed position of agent $n$ at time $t$ is represented as $\mathbf{x}_n^t = (x_n^t, y_n^t)$.  
The observation set over a time horizon $T_{obs}$ is:  
$S = \left\{ \mathbf{x}_n^t \right\}_{0:N}^{-T_{obs}:0}$.  
For autonomous planning, we define a predictor $\Phi$ that predicts future trajectories of surrounding agents:  
$\hat{F} = \Phi(S), \quad \text{where} \quad \hat{F} = \left\{ \boldsymbol{\hat{\mathbf{x}}}_n^{t} \right\}_{1:N}^{1:T_{fut}}$.  
The planner $\Theta$ generates a trajectory for the ego agent:  
$\hat{\mathbf{y}} = \hat{\mathbf{x}}_0^{1:T_{fut}} = \Theta(S_0, \hat{F})$.  
Our goal is to train $\Theta$ to produce future plans that satisfy downstream objectives modeled as a reward $R$. 

\begin{figure*}[t]
\centering
\centerline{\includegraphics[width=0.9\linewidth]{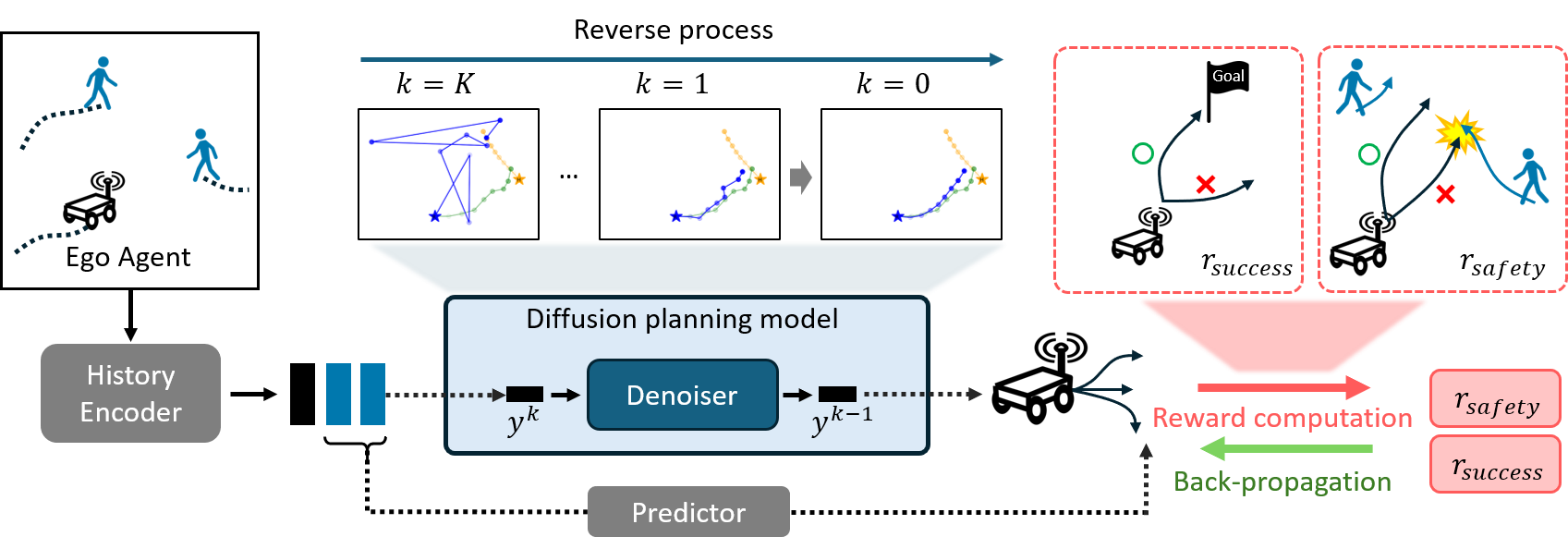}}
\caption{
Overview of the proposed method for optimizing non-differentiable rewards in diffusion-based motion planning for dynamic multi-agent environments.
During the reverse diffusion process, the model iteratively denoises at each step, storing the log likelihood at each diffusion step.
At the final step ($k=0$), safety and effectiveness-related rewards ($r_\text{safe}$, $r_\text{valid}$) are computed from generated sample.
These rewards are then used to update the denoising model by weighting the log likelihood maximization with the calculated reward, enabling direct optimization of critical autonomy objectives via reward maximization.
}
\label{fig:viz_main_method}
\vspace{-5pt}
\end{figure*}

\subsection{Diffusion Motion Planner Training}
We use a diffusion model to formulate the planning process as a denoising process~\cite{ho2020denoising}.
At diffusion timestep $k=0$, the trajectory of the ego agent is sampled as:  
\begin{equation}
    \hat{\mathbf{y}} \sim p_\theta\left(\mathbf{y}^{0:K} \right) = p\left(\mathbf{y}^K\right) \prod_{k=1}^K p_\theta\left(\mathbf{y}^{k-1} \mid \mathbf{y}^k, S_0, \hat{F} \right)
\end{equation}
Each reverse diffusion step follows a Gaussian transition:  
\begin{equation}
\label{eq:reverse_diffusion}
    p_\theta\left(\mathbf{y}^{k-1} \mid \mathbf{y}^k, S_0, \hat{F} \right) = \mathcal{N}\left(\mathbf{y}^{k-1} ; \boldsymbol{\tilde{\mu}}, \mathbf{\Sigma} \right), p\left(\mathbf{y}^K\right) \sim \mathcal{N}\left(0, \mathbf{I}\right)
\end{equation}

For conventional diffusion motion planning methods, the planner is trained by maximizing the variational lower bound:  
\begin{equation}
    \mathcal{L}_{\text{DDPM}}(\theta) = \mathbb{E} \left[\left\|\boldsymbol{\tilde{\mu}}(\mathbf{y}^0, k) - \boldsymbol{\mu}_\theta(\mathbf{y}^k, S_0, \hat{F}, k)\right\|^2 \right]
\end{equation}
where $\boldsymbol{\tilde{\mu}}$ is the posterior mean of the forward process.
However, this objective does not explicitly enforce motion planning constraints, often leading to suboptimal trajectories.
In that sense, we leverage RL to explicitly optimize a diffusion planning model on non-differentiable objectives.

\subsection{Training with Reinforcement Learning}
Using RL method in~\cite{black2024training}, we optimize diffusion-based planner on certain downstream objectives ($R$) such as safety and goal achievement as shown in Fig.~\ref{fig:viz_main_method}.
The RL objective is to maximize the reward signal $R$ defined as:  
\begin{equation}
\label{eq:rl_obj_orig}
    \mathcal{J} = \mathbb{E} \left[ R(\hat{\mathbf{y}}, \hat{F}) \right].
\end{equation}
In our formulation, the state and action each corresponds to $(\mathbf{y}^k, S_0, \hat{F})$ and $p_\theta\left(\mathbf{y}^{k-1} \mid \mathbf{y}^k, S_0, \hat{F} \right)$.
We define the denoising process from timestep $K$ to 0 as a sequential decision-making framework, modeled as a multi-step MDP:  
\begin{equation}
\label{eq:rl_obj_cum}
    \mathcal{J} = \mathbb{E} \left[ \sum_{k=0}^{K} R(\mathbf{y}^k, \hat{F}) \right].
\end{equation}
Each diffusion step in Eq. (\ref{eq:reverse_diffusion}) is regarded as probabilities of action at given state, and we aim to maximize objectives accumulated through entire diffusion steps.
Using policy gradient theorem and importance sampling estimator~\cite{kakade2002approximately}, the gradient of the objective is:  
\begin{multline}
    \nabla_\theta \mathcal{J} = \mathbb{E} \left[ \sum_{k=0}^K \frac{p_\theta\left(\mathbf{y}^{k-1} \mid \mathbf{y}^k, S_0, \hat{F} \right)}{p_{\theta_{\text{old}}}\left(\mathbf{y}^{k-1} \mid \mathbf{y}^k, S_0, \hat{F} \right)} \right. \\
    \left. \nabla_\theta \log p_\theta\left(\mathbf{y}^{k-1} \mid \mathbf{y}^k, S_0, \hat{F} \right) R(\hat{\mathbf{y}}, \hat{F}) \right].
\end{multline}
Here, $\theta_{\text{old}}$ represents the parameters of the previous planner used for importance sampling.  
To prevent variance of gradient being too large, we use the \textit{Baseline}~\cite{schulman2015high} method. 
Here, the advantage function is used to distinguish between high-quality and low-quality samples within the batch:
\begin{equation}
\label{eq:final_obj}
    \nabla_\theta \mathcal{J} = \mathbb{E} \left[ \sum_{k=0}^{K} \frac{p_\theta\left(\mathbf{y}^{k-1} \mid \mathbf{y}^k, S_0, \hat{F} \right)}{p_{\theta_{\text{old}}}\left(\mathbf{y}^{k-1} \mid \mathbf{y}^k, S_0, \hat{F} \right)} \frac{R - \bar{R}}{\sigma_R} \right]
\end{equation}


\subsection{Reward Function Selection}
\label{sec:reward_select}
To train the model for downstream objectives of autonomous planning, we define a reward for each category of safety and effectiveness.
For conventional deep learning-based training methods, the objective function must be differentiable to allow gradient-based optimization.  
Our method, on the other hand, does not directly differentiate and back-propagate the rewards.
Instead, the denoising model empirically learns to generate denoising actions that results in maximal reward with RL-based algorithm.
This enables optimization on non-differentiable rewards, such as collision rate (the ratio of collisions among agent pairs) and success rate (the ratio of agents reaching their goal), both of which are crucial towards safe and effective autonomous planning.  

To optimize multiple objectives simultaneously, we define the total reward \( R \) as a weighted sum of multiple rewards: $R = \sum_{m=1}^{M} \lambda_m r_m$.
Each reward is scaled by its importance factor \( \lambda_m \), which is empirically determined.  
Notably, our method allows incorporating an arbitrary number and type of rewards, as discussed in Sec.~\ref{sec:diverse_obj}.

\subsection{Dynamic Thresholding for Reward Shaping}

Naively applying the proposed training method to motion planning leads to a well-known policy optimization issue: learning instability under sparse rewards.
In autonomous driving, downstream objectives such as collision rate and success rate often yield sparse rewards.  
When rewards are derived from evaluation metrics, collisions may be too rare, or many agents may fail to reach their goals, making it difficult to extract meaningful rewards within a batch.
This leads to unstable training and results in suboptimal performance.

To address this, we introduce \textit{dynamic thresholding}, a method that adaptively adjusts reward sparsity at each iteration.  
Since autonomous driving rewards are often threshold-based, we dynamically adjust the threshold \( \epsilon \) to reshape rewards based on their sparsity. The rewards used—both the collision rate and the success rate—are defined as the proportion of cases where the Euclidean distance between the ego agent and surrounding agents, and between the ego agent and the goal, respectively, fall below a predefined threshold.
Each batch consists of \( B \) agents, where rewards are computed at the agent level.
For example, in the case of the success rate reward, if all agents take optimal actions, the total reward is \( B \), whereas if all fail, it is \( 0 \).  
To prevent reward collapse to either extreme, we regulate \( \epsilon \) such that the mean reward over the batch remains close to \( B/2 \), ensuring stable learning.

In more detail, as iterations progress, we incrementally adjust the threshold by using adaptation rate $\alpha$ that decreases over iterations, followed by recomputing the reward. 
The direction of the threshold adjustment is also crucial to bringing the reward closer to $B/2$. 
Therefore, the newly computed mean reward must be closer to $B/2$; if it moves further away, we reverse the sign of the adaptation rate. Considering the trade-off between model performance and training time, and to prevent the reward adjustment from remaining incomplete due to the reward mean not approaching $B/2$ within deviation $\delta$, we stop adjusting the threshold after a certain number of iterations $J$, thereby maintaining training time. 
The final shaped reward is used in Eq. (\ref{eq:final_obj}), driving the policy gradient updates for the model.  
The detailed procedure is described in Algorithm~\ref{alg:alg1}.

\begin{algorithm}[t]
\DontPrintSemicolon
\caption{Dynamic thresholding algorithm}\label{alg:alg1}

\KwData{Thresholding max iter. $J$, batch size $B$, init. threshold $\epsilon_{init}$, deviation $\delta$, adapt. rate $\alpha$}

\For{$m \in \{col. rate, succ. rate, etc\}$}
{
    $\epsilon \gets \epsilon_{m, init}$ \\
    \For{$j=1 \textbf{ to } J$}{
    
        Compute reward: $\boldsymbol{r_{m}({\epsilon})} = [r_{m}^1({\epsilon}), r_{m}^2({\epsilon}), \dots, r_{m}^B({\epsilon})]$ 

        \tcp{\small Reward Adjusting Process}
        
        \If{$\left| \sum \boldsymbol{r_m({\epsilon})} - B/2 \right| \leq \delta$}
            {
                \textbf{break}
            }
        \Else{
            $\epsilon_{new} = \epsilon + \mathcal{\alpha}(j) \cdot \epsilon$ \\ 
            \If{$\left| \sum \boldsymbol{r_m({\epsilon_{new})}} - B/2 \right| > \left| \sum \boldsymbol{r_m({\epsilon})} - B/2 \right|$}{
                $\epsilon_{new} = \epsilon - \mathcal{\alpha}(j) \cdot \epsilon$ \\
            }
            $\epsilon \gets \epsilon_{new}$
            } 
            
        }    
}
\tcp{\small Multi-Reward Process} 
$R = \sum \lambda_m r_m(\epsilon)$, Using $R$ to Eq. (\ref{eq:final_obj}). \\
\end{algorithm}
\section{EXPERIMENTS}

\subsection{Datasets}   
We validate the effectiveness of our method on two pedestrian trajectory datasets: CrowdNav~\cite{chen2019crowd} and ETH-UCY~\cite{pellegrini2009you, lerner2007crowds}. 
For CrowdNav dataset, past and future trajectory timespans are both 3.2s with 2.5 FPS.
Each CrowdNav scene contains one designated ego agent and five surrounding agents.
For ETH-UCY dataset, the past and future trajectory timespans are respectively set to 3.2 and 4.8 seconds with 2.5 FPS. 
We use ETH scene as validation set and others (Hotel, Zara1, Zara2, Univ) as training set.
Following previous work~\cite{kedia2023game}, one individual in each scene is designated as an ego agent and others are treated as surrounding agents. 

\subsection{Metrics}
Autonomous systems must generate safe trajectories that reach the destination while avoiding collisions with the surroundings.  
To evaluate the planner's performance based on these criteria, we employ two primary metrics: collision rate and success rate, both of which are also used as rewards.  
Additionally, we use Average Displacement Error (ADE), a metric commonly used in motion planning tasks.  
However, it is important to note that non-differentiable metrics, such as collision rate and success rate, are more relevant in real-world applications.  
Unlike ADE, which measures the proximity of the generated trajectory to the ground truth, these non-differentiable metrics evaluate whether the planned motion satisfies fundamental constraints, making them more practical for deployment.  
The initial thresholds used to compute collision rate and success rate are based on prior studies~\cite{kedia2023game, yue2022human} for each dataset: 0.6 m, 0.2 m for CrowdNav and 0.2 m, 0.5 m for ETH-UCY.

\begin{table*}[]
\centering
\caption{Comparison of motion planning performance across the critical metrics in autonomous systems - safety and effectiveness - as well as widely used conventional planning metric, ADE. 
Best performance in \textbf{bold}.}
\label{tab:table1_main}
\renewcommand{\arraystretch}{1.2}
\begin{tabular}{cl|ccc|ccc}
\toprule
\multicolumn{2}{c|}{Dataset}                                                      & \multicolumn{3}{c|}{CrowdNav}                                              & \multicolumn{3}{c}{ETH-UCY}                                               \\ \hline
\multicolumn{2}{c|}{\multirow{2}{*}{Metric}}                                      & Safety                   & Effectiveness           & Conventional          & Safety                   & Effectiveness           & Conventional         \\
\multicolumn{2}{c|}{}                                                             & (col. rate $\downarrow$) & (succ. rate $\uparrow$) & (ADE $\downarrow$)    & (col. rate $\downarrow$) & (succ. rate $\uparrow$) & (ADE $\downarrow$)   \\ \hline
\multicolumn{1}{c|}{\multirow{6}{*}{\rotatebox{90}{Method}}} & DTPP~\cite{huang2024dtpp}                               & 0.0669                   & 0.3853                  & 0.2354                & 0.1129                   & 0.3473                  & 0.5184               \\
\multicolumn{1}{c|}{}                        & DIPP~\cite{huang2023differentiable} & 0.0446                   & 0.4004                  & 0.2262                & 0.1022                   & 0.3834                  & 0.4993               \\
\multicolumn{1}{c|}{}                        & GameTheoretic~\cite{kedia2023game}                      & 0.0276                   & 0.4418                  & 0.1991                & 0.0906                   & 0.3353                  & 0.4769               \\ \cline{2-8} 
\multicolumn{1}{c|}{}                        & Backbone                           & 0.0628                   & 0.3540                  & 0.2663                & 0.0965                   & 0.3635                  & 0.5000               \\
\multicolumn{1}{c|}{}                        &  Backbone + \cite{dhariwal2021diffusion}                                  & 0.0562     & 0.3656    & 0.2590 & 0.0906     & 0.3368    & 0.5406 \\
\multicolumn{1}{c|}{}                        & Backbone + Ours                            & \textbf{0.0150}          & \textbf{0.4931}         & \textbf{0.1984}       & \textbf{0.0879}          & \textbf{0.3910}         & \textbf{0.4683}      \\ \bottomrule
\end{tabular}
\vspace{-5pt}
\end{table*}

\subsection{Baselines}

We compare our method with the following baselines:

\noindent\textbf{GameTheoretic}~\cite{kedia2023game} enhances safety by predicting adversarial trajectories for surrounding agents, allowing the planner to proactively respond to rare but critical scenarios.

\noindent\textbf{DIPP}~\cite{huang2023differentiable} optimizes planning objectives by leveraging predicted trajectories of surrounding agents.  
It formulates the optimization target as a differentiable function, enabling joint learning of both the planner and forecaster.

\noindent\textbf{DTPP}~\cite{huang2024dtpp} is a state-of-the-art autonomous planner that employs a tree-structured policy planner and a query-centric transformer.  
By introducing a differentiable joint training framework, DTPP improves planning quality through integrated learning and effectively captures interactions between prediction and planning.

\noindent\textbf{Gradient guidance}~\cite{dhariwal2021diffusion} is a sampling technique applied to pretrained diffusion models to generate samples conditioned on differentiable metrics.  
The gradient of the optimization metric is incorporated as an additional denoising signal.  
For motion planning, we use the collision cost~\cite{kedia2023game} as the additional guidance source.

Our method focuses solely on training the planner without updating the forecaster.  
For a fair comparison, we pre-train the forecaster in the baselines and keep it frozen while training the planner, ensuring a direct performance comparison with ours.
All baseline models are conventionally trained with gradients from L2 with GT trajectory and collision cost.

\subsection{Implementation Details}
\noindent\textbf{Backbone Model.}
We build a diffusion-based motion planner based on prior work using the CrowdNav and ETH-UCY datasets.  
For CrowdNav, we integrate GameTheoretic~\cite{kedia2023game} as the forecaster and adapt the MID~\cite{gu2022stochasticMID} decoder—originally designed for trajectory prediction—as the planner.  
The MID decoder receives encodings from the GameTheoretic encoder.  
For ETH-UCY, we employ diffusion models for both the forecaster and planner, utilizing MID in both components.  
We model agent interactions using attention over multimodal forecasts, an LSTM for trajectory embeddings, and a Graph Transformer for ego-centered representations.  
The reason for using different backbones for each dataset is to conduct experiments on existing diffusion model architectures that are specialized for the complexity of each dataset.
The backbone model is initially trained without downstream objectives, serving as a baseline to evaluate the impact of reward-based fine-tuning.

\noindent\textbf{Reward-based training details.}
We train our model using a mixture of rewards derived from downstream metrics.  
For CrowdNav, we use collision rate, success rate, and discomfort rate as components of rewards, with importance factors $\lambda_{m}$ respectively set to 4, 5, and 1. For ETH-UCY, collision rate and success rate are used with $\lambda_{m}$ set as 3 and 7.
During training, the batch size determines the number of samples used per reward calculation.  
We use 128 samples per batch for CrowdNav and 8 samples per batch for ETH-UCY.
\section{RESULTS}

\begin{figure*}[tb]
    \centering
    \includegraphics[width=0.95\textwidth]{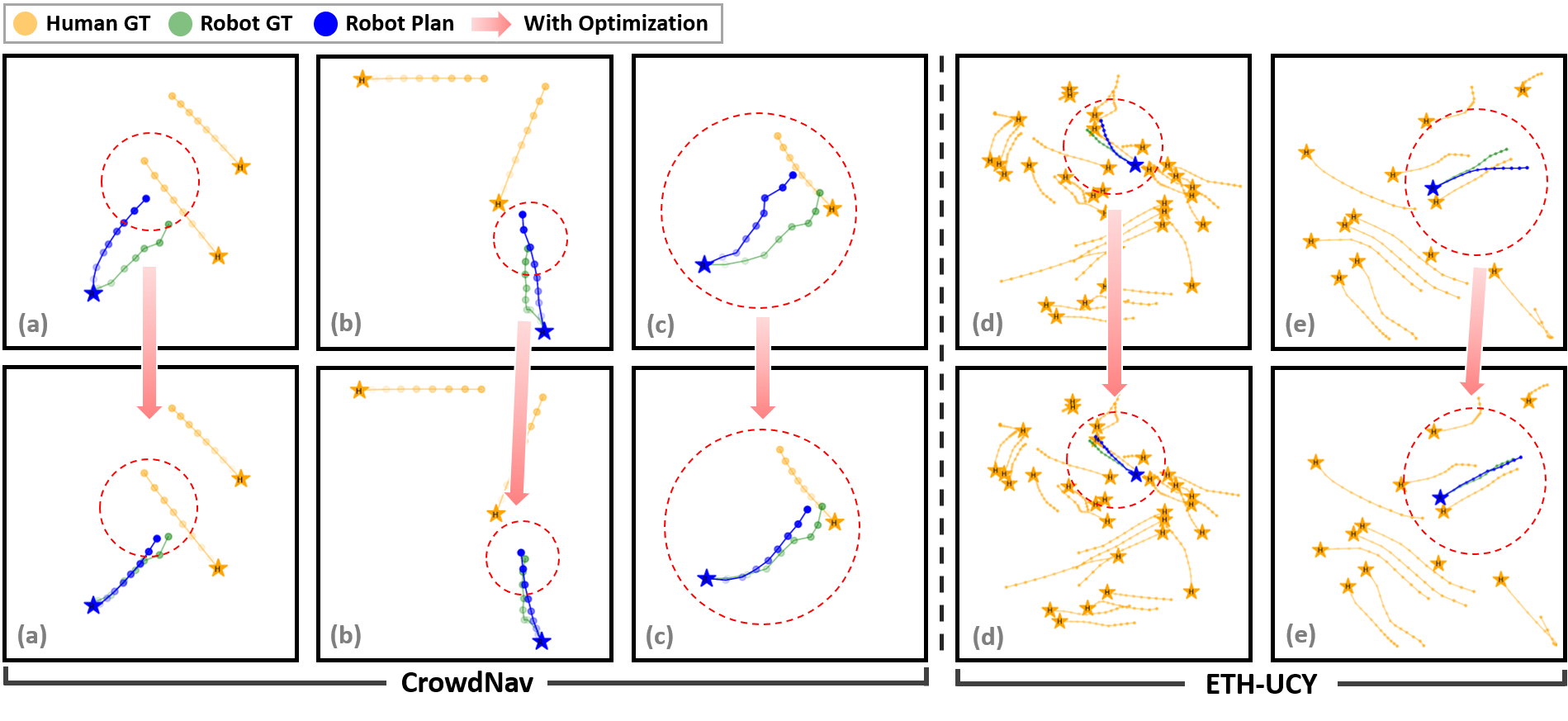}
    \caption{
    Qualitative results on the CrowdNav and ETH-UCY, where human (orange) represent surrounding agents and the robot (blue, green) represents the ego agent.  
    The proposed method significantly improves the backbone model in terms of (a) safety, (b) effectiveness, and (c) overall performance.  
    In ETH-UCY (d) and (e), it demonstrates strong performance even in complex, crowded scenes.
    }
    \label{fig:Figure3_qual}
    \vspace{-7.5pt}
\end{figure*}
%

\begin{table}[t]
\centering
\footnotesize
\caption{
    Ablation study on optimizing with single reward with CrowdNav dataset.
    Optimizing each specific reward (col. rate, succ. rate, discomf. rate) results in the corresponding metric achieving the best performance.
    Optimizing on multi-reward $R$ results in comparable performance on all metrics. 
}
\resizebox{\columnwidth}{!}{%
\begin{tabular}{c|ccc}
\hline
\multirow{2}{*}{\begin{tabular}[c]{@{}c@{}}Optimization\\ reward\end{tabular}} & \multicolumn{3}{c}{Metrics} \\ \cline{2-4} 
 & \multicolumn{1}{l}{Collision rate$\downarrow$} & \multicolumn{1}{l}{Success rate$\uparrow$} & \multicolumn{1}{l}{Discomfort rate$\downarrow$} \\ \hline
Collision rate & \textbf{0.0146} & 0.4493 & 0.3445 \\
Success rate & \textbf{0.0146} & \textbf{0.4987} & 0.3498 \\
Discomfort rate & 0.0191 & 0.4832 & \textbf{0.2681} \\ \hline
Multi-reward $R$ & 0.0150 & 0.4931 & 0.3178 \\ \hline
\end{tabular}
}
\label{tab:desiredmetricoptimization}
\vspace{-7pt}
\end{table}

\subsection{Main Results}
\noindent\textbf{Quantitative Results.}
As shown in Tab.~\ref{tab:table1_main}, training the backbone model with our method achieves the best overall performance across all metrics.  
In particular, the effectiveness of our approach is most evidently manifested with the noticeable improvement on collision rate metric.
The backbone diffusion model solely trained on ground truth (GT) trajectories does not leverage collision information during training, leading to a high collision rate.  
However, when trained with our multi-reward framework, which includes collision rate as a key objective, the model achieves a significant reduction in collision—from 0.0628 to 0.0150 on the CrowdNav.  
This also surpasses baselines that rely on collision cost function, demonstrating the effectiveness of directly optimizing non-differentiable collision rate over indirect gradient-based optimization via collision cost.

Furthermore, our approach outperforms gradient-based sampling method~\cite{dhariwal2021diffusion}, reaffirming the effectiveness of optimizing non-differentiable metrics through reward-driven training.  
In detail, we implement \cite{dhariwal2021diffusion} by leveraging the gradient from collision loss from~\cite{kedia2023game}.
In addition, our method achieves state-of-the-art performance not only for collision rate but also for success rate and ADE, showcasing the versatility of our approach.
By utilizing non-differentiable rewards for diverse categories that lack differentiable metrics, our approach generates safer motions while effectively guiding the agent toward its goal.

\noindent\textbf{Qualitative Results.}
By incorporating rewards from multiple metrics, the model is fine-tuned to generate more effective plans that adapt to diverse scenarios, as demonstrated in the qualitative results.  
In Fig.~\ref{fig:Figure3_qual} (a), the original model generates a trajectory that risks colliding with surrounding agents.  
Here, the collision rate reward plays a crucial role in guiding the model to learn safer planning behavior, effectively reducing the likelihood of collisions.  
Similarly, in Fig.~\ref{fig:Figure3_qual} (b), the success rate is significantly improved, demonstrating the model's enhanced ability to reach its goal.  
In Fig.~\ref{fig:Figure3_qual} (c), the initial plan exhibits poor performance both on collision and success.
In comparison, the mixture reward effectively optimizes all aspects, enabling the model to generate a well-balanced trajectory that comprehensively addresses these challenges.
In Fig.~\ref{fig:Figure3_qual} (d) and (e), our methodology demonstrates improvement in performance of downstream tasks even in scenes with high density.

\begin{figure}[t]

  \centering
   \includegraphics[width=0.9\linewidth]{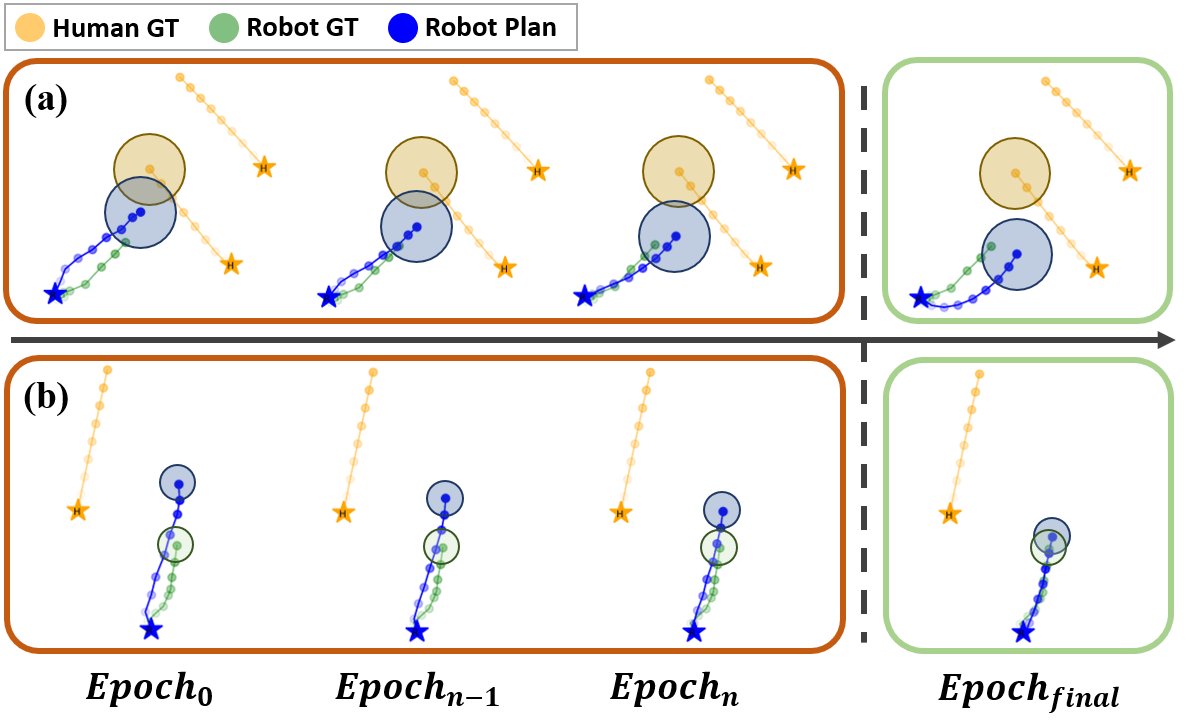}

   \caption{
    Improvement of planning results over training epochs on CrowdNav using the proposed method, from left to right.
    The top row (a) shows how our method generates a collision-free trajectory. Each circle represents the threshold for collision.
    The bottom row (b) shows the improvement in goal-reaching over the same scene. Each circle represents the threshold goal reaching.
   }
   \label{fig:desiredmetricoptimizationfig}
   \vspace{-7.5pt}
\end{figure}

\subsection{Ablation Studies}

\noindent\textbf{Singular Optimization on Diverse Objectives.}
\label{sec:diverse_obj}
While safety and effectiveness are critical objectives in motion planning, other factors, such as comfort, also play a crucial role in ensuring high-quality trajectories.  
Specifically, the discomfort rate measures the proportion of trajectories with a jerk exceeding a predefined threshold, indicating that excessive acceleration changes lead to discomfort.  
Table~\ref{tab:desiredmetricoptimization} presents the individual optimization results across different metrics (upper three rows).  
To examine whether reward-based learning enables the model to specialize in the metric used for training, we evaluated models optimized with a single reward.  
Since motion planning metrics are highly interdependent, optimizing a single metric often enhances overall performance across multiple metrics.
For instance, training the model on success rate results in endpoints closer to the GT, which in turn decreases the collision rate.  

In addition to such holistic improvement, we also observe a significant improvement in the corresponding metric.
This trend is also illustrated in Fig.~\ref{fig:desiredmetricoptimizationfig}, where planning results evolve as our training method is applied.  
Figure~\ref{fig:desiredmetricoptimizationfig} (a) shows the outcome when optimizing for collision rate alone, while Fig.~\ref{fig:desiredmetricoptimizationfig} (b) presents the result of training solely with success rate.  
These findings demonstrate that our approach effectively fine-tunes the model to optimize a specific desired metric.  

Furthermore, unlike conventional methods such as GameTheoretic~\cite{kedia2023game} which primarily focus on a single safety objective, our approach enables a simultaneous optimization of multiple rewards.
The multi-reward result in Tab.~\ref{tab:desiredmetricoptimization} (lower row) shows that optimizing with multiple rewards yields strong performance across all accompanied metrics.
This flexibility highlights the broader applicability of our approach, making it suitable for any specific motion planning scenarios.

\begin{table*}[ht!]
\caption{
    Comparison of single reward-based optimization performance between \textbf{\textcolor{Rhodamine}{non-differentiable}} and conventional \textbf{\textcolor{Turquoise}{differentiable}} objectives in CrowNav dataset.
    Optimizing with non-differentiable rewards outperforms optimizing differentiable metrics in their respective categories. 
    }
\renewcommand{\arraystretch}{1.1}
\setlength\tabcolsep{12.0pt}
\resizebox{\textwidth}{!}{%
\begin{tabular}{cc|ccc|c|cc}
\hline
\multicolumn{2}{c|}{}                                                                                                                 & \multicolumn{3}{c|}{Effectiveness}                                              & Safety                              & \multicolumn{2}{c}{Comfort}       \\
\multicolumn{2}{c|}{\multirow{-2}{*}{Reward}}                                                                                         & \cellcolor[HTML]{FFFFFF}(\textbf{\textcolor{Turquoise}{ADE}} $\downarrow$)& \cellcolor[HTML]{FFFFFF}(\textbf{\textcolor{Turquoise}{FDE}} $\downarrow$)& (\textbf{\textcolor{Rhodamine}{Success rate}} $\uparrow$)& \cellcolor[HTML]{FFFFFF}(\textbf{\textcolor{Rhodamine}{Collision rate}} $\downarrow$)& (\textbf{\textcolor{Turquoise}{Jerk}} $\downarrow$)& (\textbf{\textcolor{Rhodamine}{Discomfort rate}} $\downarrow$)\\ \hline
\multicolumn{1}{c|}{}                                                        & \cellcolor[HTML]{FFFFFF}{\color[HTML]{1F1F1F} \textbf{\textcolor{Turquoise}{ADE}}}     & 0.2089                        & 0.2953                        & 0.4437          & 0.0171                              & 0.5473          & 0.3510           \\
\multicolumn{1}{c|}{}                                                        & \textbf{\textcolor{Turquoise}{FDE}}                                                    & 0.2109                        & 0.2976                        & 0.4377          & 0.0171                              & 0.5579          & 0.3613          \\
\multicolumn{1}{c|}{\multirow{-3}{*}{{Effectiveness}}}                   & \textbf{\textcolor{Rhodamine}{Success rate}}& \textbf{0.1982}               & \textbf{0.2791}               & \textbf{0.4987} & \textbf{0.0146}                     & \textbf{0.5136} & \textbf{0.3498} \\ \hline
\multicolumn{1}{c|}{\cellcolor[HTML]{FFFFFF}}                                & \cellcolor[HTML]{FFFFFF}{\color[HTML]{1F1F1F} \textbf{\textcolor{Turquoise}{Collision cost}}}& 0.2161& 0.3057& 0.4301& 0.0179& 0.5984& 0.3861\\
\multicolumn{1}{c|}{\multirow{-2}{*}{\cellcolor[HTML]{FFFFFF}{Safety}}}  & \textbf{\textcolor{Rhodamine}{Collision rate}}& \textbf{0.2128}& \textbf{0.2969}& \textbf{0.4493}& \textbf{0.0146}& \textbf{0.5382}& \textbf{0.3445}\\ \hline
\multicolumn{1}{c|}{\cellcolor[HTML]{FFFFFF}}                                & \textbf{\textcolor{Turquoise}{Jerk}}& 0.1995                        & 0.2846                        & 0.4832          & 0.0203                              & 0.4800            & 0.2783          \\
\multicolumn{1}{c|}{\multirow{-2}{*}{\cellcolor[HTML]{FFFFFF}{Comfort}}} & \textbf{\textcolor{Rhodamine}{Discomfort rate}}& \textbf{0.1985}               & \textbf{0.2829}               & \textbf{0.4877} & \textbf{0.0191}                     & \textbf{0.4741} & \textbf{0.2681} \\ \hline
\end{tabular}%
}
\vspace{-5pt}
\label{tab:differentiable_reward_ablation}
\end{table*}

\noindent\textbf{Comparison with Differentiable Reward Optimization.}  
Unlike diffusion models, other deep learning-based approaches are constrained to using only differentiable functions during training.  
To optimize non-differentiable metrics, these models typically rely on surrogate loss functions that approximate the desired objectives.  
We propose a reward-based approach that enables diffusion models to learn downstream objectives while directly incorporating non-differentiable metrics into training.  
To show that any reward including non-differentiable metrics can be effectively utilized, we compare performances under two conditions.  
First, we use rewards derived from differentiable metrics or loss functions commonly used in training.  
Second, we employ discrete, non-differentiable metrics as direct rewards.  

For each downstream objective, the differentiable metrics are defined as follows: ADE and FDE (Final Displacement Error) for Effectiveness, the collision cost loss function from GameTheoretic~\cite{kedia2023game} for Safety, and Jerk for Comfort.  
As shown in Tab.~\ref{tab:differentiable_reward_ablation}, the experimental results indicate that differentiable metrics also can serve as reward functions that improve model performance.  
However, using non-differentiable, discrete metrics as rewards generally leads to higher overall performance compared to their differentiable counterparts.  
These findings confirm that our model can be effectively optimized with various reward functions, while particularly achieving superior performance when trained with non-differentiable metrics.

\begin{figure}[t]
  \centering

   \includegraphics[width=\linewidth]{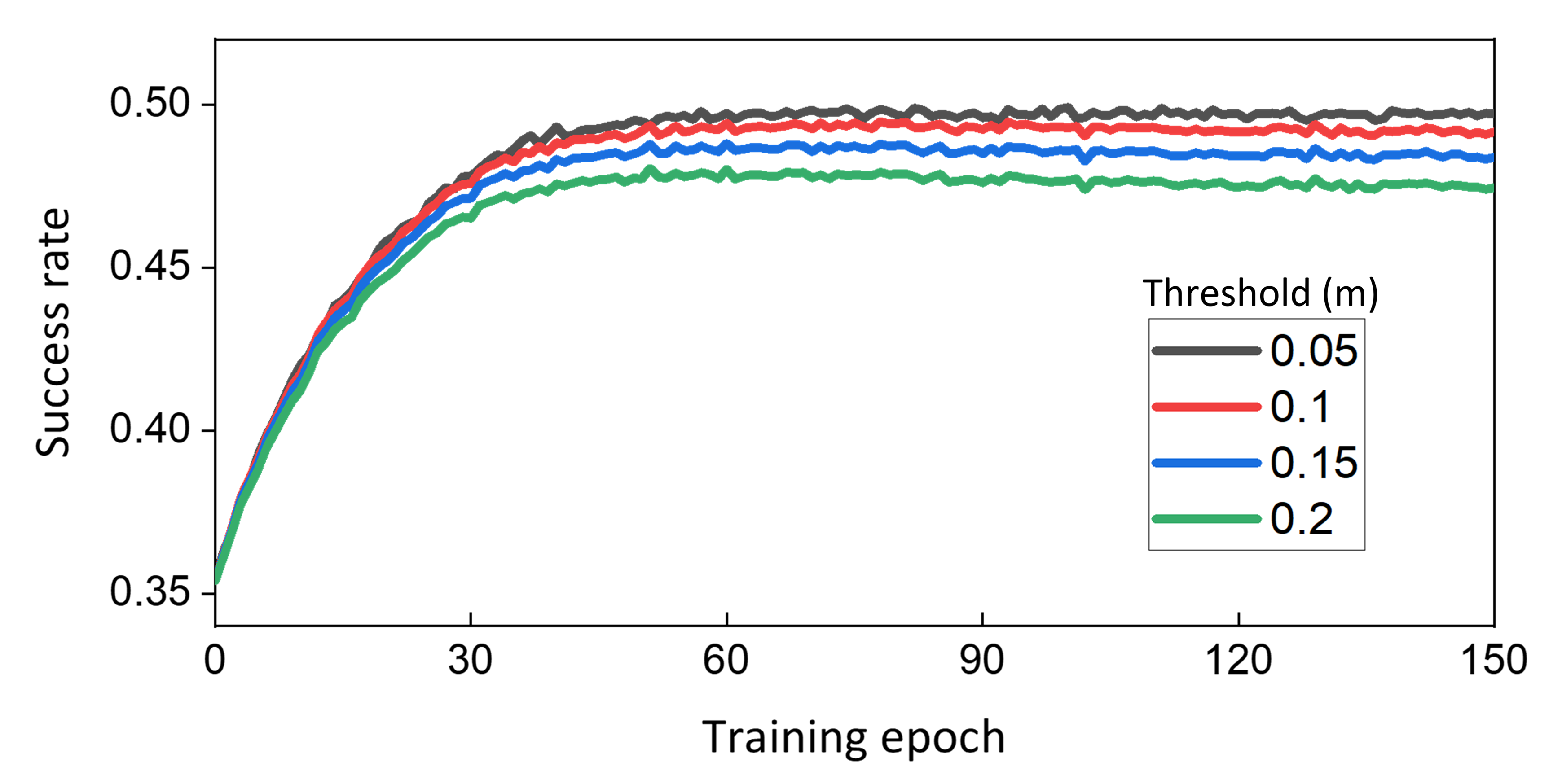}

   \caption{
    Comparison of optimization performances with different success rate thresholds.
    Adjusting the threshold impacts the success rate achieved over training epochs, emphasizing the importance of setting the threshold to maximize the information conveyed by the reward.
    This shows the importance of actively adjusting the threshold in our method to improve optimization effectiveness.
   }
     \label{fig:ablation_threshold}
     \vspace{-7pt}
\end{figure}

The superior performance of non-differentiable rewards can be attributed to their discrete nature, which differs from smooth, differentiable reward functions.  
Unlike differentiable rewards which provide continuous gradients, discrete rewards are influenced by threshold settings.  
Stricter thresholds during training create a clearer distinction between good and poor samples within a batch, leading to more effective optimization.  
We verify this phenomenon by observing that, for single-reward optimization, performance decreases as the reward threshold becomes less sensitive as shown in Fig.~\ref{fig:ablation_threshold}.  
Nevertheless, the success rate still surpasses the best performance among baseline methods, even in the worst-case scenario ($th=0.2$).

\noindent\textbf{Dynamic Thresholding.}  
We conduct an ablation study to evaluate the effect of dynamic thresholding, as shown in Tab.~\ref{tab:my-table_4}.
To ensure a precise comparison, we perform experiments on the safety downstream task leveraging collision cost—a differentiable metric where dynamic thresholding can be applied.
We compare collision rate and collision cost as reward, both with and without dynamic thresholding.
Experimental results show that applying dynamic thresholding to both rewards consistently improves collision rate performance.
These findings confirm the effectiveness of dynamic thresholding, demonstrating greater improvements when applied to non-differentiable rewards.

\begin{table}[t!]
\caption{Ablation results on dynamic thresholding  (DT). The dynamic thresholding algorithm is effective for both collision rate (non-differentiable) and collision cost (differentiable). Additionally, non-differentiable (collision rate) with DT applied achieves the highest performance.}
\resizebox{\columnwidth}{!}{%
\begin{tabular}{c|ccc}
\hline
Reward & \begin{tabular}[c]{@{}c@{}}Safet\\ (Collision rate)\end{tabular} & \begin{tabular}[c]{@{}c@{}}Effectiveness\\ (Success rate)\end{tabular} & \begin{tabular}[c]{@{}c@{}}Conventional\\ (ADE)\end{tabular} \\ \hline
Collision cost (w/o DT) & 0.0216 & 0.431 & 0.2165 \\
Collision cost (w/ DT) & 0.0179 & 0.4301 & 0.2161 \\
Collision rate (w/o DT) & 0.0232 & 0.4301 & 0.2173 \\
Collision rate (w/ DT) & \textbf{0.0146} & \textbf{0.4493} & \textbf{0.2128} \\ \hline
\end{tabular}
\vspace{-18.5pt}
}
\label{tab:my-table_4}
\end{table}

\section*{ACKNOWLEDGMENT}
This work was supported by the Institute of Information \& Communications Technology Planning \& Evaluation(IITP) grant funded by the Korea government(MSIT) (No. RS-2025-02219277, AI Star Fellowship Support Project(DGIST)), Korean ARPA-H Project through the Korea Health Industry Development Institute (KHIDI), funded by the Ministry of Health \& Welfare, Republic of Korea (No. RS-2024-12345678), and by the National Research Foundation of Korea(NRF) grant funded by the Korea government(MSIT) (NRF2022R1A2B5B03002636).




\bibliographystyle{IEEEtran}
\bibliography{IEEEabrv,mybib}

\end{document}